\documentclass[journal,twocolumn,letterpaper]{IEEEJERM}

\ifCLASSINFOpdf
\else
\fi

\usepackage{times,amsmath,epsfig}
\usepackage{fancyhdr}
\usepackage{amsmath}
\usepackage{amsfonts}
\usepackage{amssymb}
\usepackage{array}
\usepackage{graphicx}
\usepackage{url}
\usepackage{bm}
\usepackage{breqn}
\usepackage{soul}
\usepackage{amssymb}
\usepackage[breaklinks=true]{hyperref}
\usepackage{breakcites}
\usepackage{xspace}
\usepackage{cite}
\usepackage{multirow}
\usepackage{mathtools} 
\usepackage{dsfont} 
\usepackage{array}
\usepackage{makecell}
\usepackage{verbatim}
\usepackage{lipsum}
\usepackage[table,xcdraw]{xcolor}
\usepackage[caption=false]{subfig}

\usepackage{etoolbox}
\makeatletter
\patchcmd{\@makecaption}
  {\scshape}
  {}
  {}
  {}
\makeatletter
\patchcmd{\@makecaption}
  {\\}
  {.\ }
  {}
  {}
\makeatother
\def\tablename{Table}
\newcommand*{\mean}[1]{\bar{#1}}

\newcommand*{\humandata}{\mathbf{d}} 

\newcommand*{\phase}{\varphi} 
\newcommand*{\task}{\chi} 
\newcommand*{\subject}{\eta} 

\newcommand*{\contribution}{C}

\begin{document}

%

\vspace{1 cm}

\title{A Clinical Tuning Framework for Continuous Kinematic and Impedance Control of a Powered Knee-Ankle Prosthesis}

\author{Emma Reznick$^{1}$, T. Kevin Best$^1$ \IEEEmembership{Student Member, IEEE}, and Robert D. Gregg$^1$, \IEEEmembership{Senior Member, IEEE}
} 

\markboth{IEEE Journal of Translational Engineering in Health and Medicine}{}

\twocolumn[
\begin{@twocolumnfalse}
  
\maketitle


\begin{abstract}
 
Objective: \textnormal{Configuring a prosthetic leg is an integral part of the fitting process, but the personalization of a multi-modal powered knee-ankle prosthesis is often too complex to realize in a clinical environment. This paper develops both the technical means to individualize a hybrid kinematic-impedance controller for variable-incline walking and sit-stand transitions, and an intuitive Clinical Tuning Interface (CTI) that allows prosthetists to directly modify the controller behavior.} Methods and procedures: \textnormal{Utilizing an established method for predicting kinematic gait individuality alongside a new parallel approach for kinetic individuality, we personalize continuous-phase/task models of joint impedance (during stance) and kinematics (during swing) using tuned characteristics exclusively from level-ground walking. To take advantage of this method, we developed a CTI that translates common clinical tuning parameters into model adjustments for the walking and sit-stand controllers. We then conducted a case study where a prosthetist iteratively tuned the powered prosthesis to an above-knee amputee participant in a simulated clinical session involving sit-stand transitions and level walking, \textcolor{black}{from which incline/decline walking features were automatically calibrated}.} Results: \textnormal{The prosthetist fully tuned the multi-activity prosthesis controller in under 20 min. Each iteration of tuning (i.e., observation, parameter adjustment, and model reprocessing) took 2 min on average for walking and 1 min on average for sit-stand. The tuned behavior changes were appropriately manifested in the commanded prosthesis torques, both at the \textcolor{black}{manually tuned tasks and automatically tuned tasks (inclines)}.} Conclusion: \textnormal{The CTI leveraged able-bodied trends to efficiently personalize a wide array of walking tasks and sit-stand transitions, demonstrating the efficiency necessary for powered knee-ankle prostheses to become clinically viable.} 

Clinical impact: \textnormal{This paper introduces a clinical tuning interface that simplifies the tuning process for multimodal robotic prosthetic legs, reducing the time required from several hours to just 20 minutes thus improving clinical feasibility.}  \end{abstract}


\end{@twocolumnfalse}]

{
  \renewcommand{\thefootnote}{}%
  \footnotetext[1]{$^1$Department of Robotics at the University of Michigan, Ann Arbor, MI 48109, USA. Corresponding email: {\tt\footnotesize rdgregg@umich.edu}.}
}

%
\IEEEpeerreviewmaketitle

\section{Introduction}\label{Sec:5Intro}
\IEEEPARstart{P}{owered} prosthetic legs have the potential to restore normative, symmetrical motion to lower-limb amputees, but they are not very common, in part because of the high technical burden placed on the clinical team \cite{Gardinier2018, Huang2021, Quintero2018_CCI}. The clinical training of a prosthetist is multi-faceted, covering socket fabrication, component selection, physiology, gait analysis, and prosthesis personalization, but it is notably not an engineering degree \cite{Berger2012}. Conversely, the researchers designing emergent powered prostheses have expertise in biomechanics, robotics, and control, but often lack the clinical background to meaningfully configure those devices to patients. Much of this gap between development and implementation relates to the complexity of individualizing the motion for a specific user. This study therefore develops clinically accessible tools and methods to individualize robotic prosthetic legs.

Microprocessor-controlled, variable-damping prosthetic knees (e.g., the C-Leg from Ottobock) are personalized by adjusting or `tuning' the damping behavior of the knee with clinically-relevant, digital parameters, such as `Swing Flexion Angle'. With powered prosthetic legs, however, one or two joints are replaced with motors capable of more versatile behavior, which adds complexity to the individualization. The controllers that generate commands for these motors must generate different behaviors for different ambulation activities \cite{Reznick2020, culver2018} and often have many tunable parameters. Configuring these complex controllers for an individual's needs is often time-intensive and prohibitively difficult for clinicians \cite{Finucane2022,Gehlhar2023,Simon2014}. 
For context, powered prosthetic legs often use impedance controllers to emulate biological joints by changing the stiffness, damping, and equilibrium parameters at specific points in the gait cycle according to a finite-state machine (FSM) \cite{Sup2008, Gehlhar2023}. Simple spring-like controllers parameterize each state by defining a torque, $\tau$, as a function of joint angle, $\theta$, and angular velocity, $\dot{\theta}$:
\begin{equation}
    \tau = -K(\theta - \theta_{eq}) - B\dot{\theta}
\end{equation}
where $K$, $\theta_{eq}$, and $B$ are state-dependent tunable parameters that define the stiffness, equilibrium angle, and damping, respectively. The gait cycle is often broken into 3–5 states governed by switching rules, and distinct parameters must be defined for each possible activity (e.g., walking vs. ramp ascent), further compounding the number of tunable parameters for each individual \cite{Simon2014, Gehlhar2023}. 
\begin{figure*}[h]
    \centering
    \includegraphics[width=\textwidth]{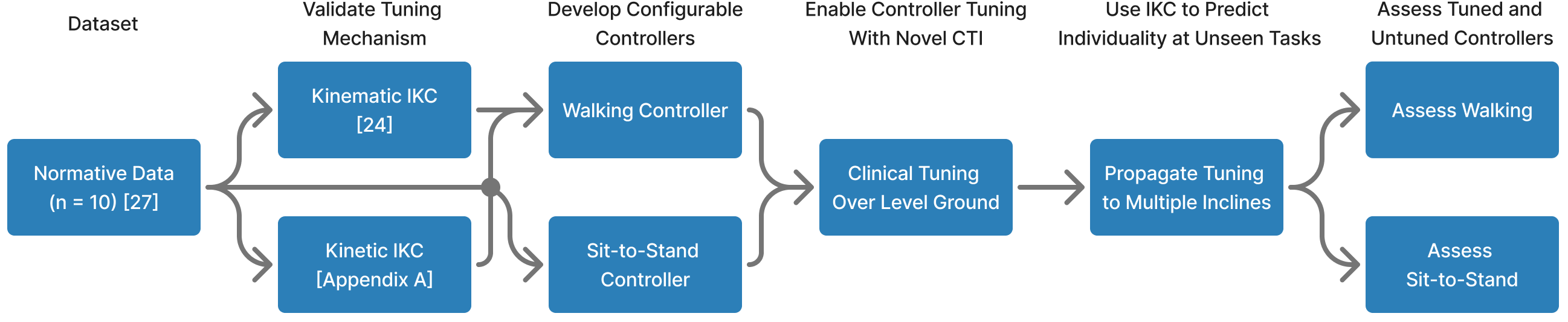}\vspace{-2mm}
    \caption{\textcolor{black}{Flowchart of the clinical tuning framework presented in this article. A prior and new analysis of able-bodied data validate the paradigm of individual kinematic/kinetic contributions (IKC) to gait. Following this validation, configurable variables are introduced to two existing hybrid kinematic-impedance controllers---one for variable-incline walking and one for sit-to-stand. A novel clinical tuning interface (CTI) is then introduced to enable manual tuning of the level-ground walking and sit-stand activities, followed by automatic tuning of incline/decline walking using the level-ground controller's kinematic and kinetic IKC. Finally, both tuned and untuned controllers are assessed and compared during sit-stand and walking activities, where the tuned incline controllers were automatically extrapolated from the manually-tuned level-ground controller.}}
    \label{fig:flowchart}
\end{figure*}
Efforts to address tuning complexity can be categorized into three main approaches: the development of tools to assist clinicians in tuning complex controllers and devices, the development of methods for real-time user tuning, and automated tuning methods that do not incorporate direct user or clinical feedback. On a parallel path, researchers are investigating the specificity and utility of preference in assistive device control and in the tuning of powered prosthetic legs, which is a major component of user and clinical feedback \cite{Alili2023, Clites2021, Ingraham2022, Shepherd2020}. Simon et al.~\cite{Simon2013} heuristically tuned a powered knee-ankle prosthesis by determining synergies within the 140 different parameters of a multimodal FSM, but the tuning process still took $\sim$5 hours to complete. Welker et al.~allowed the user to directly tune their own impedance parameters to reduce the metabolic cost of walking, with mixed success \cite{Welker2021}. Lastly, human-in-the-loop (HITL) reinforcement learning methods \cite{Alili2021, Alili2023, hong2023, wen2019, Gao2020} algorithmically tune impedance parameters based on aspects of the user's gait while walking on a treadmill. \textcolor{black}{While HITL results are overall encouraging, they do not allow prosthetist or patient input to the tuning process. Further,} these studies focus on tuning only one joint during level-ground walking and do not approach the complexity of coordinated joint motion over many tasks. Further, HITL tuning does not have an avenue to incorporate clinical expertise.

\textcolor{black}{To avoid the complexity of tuning a multimodal FSM, some groups are moving away from the FSM paradigm and using convex optimization to solve for continuous impedance parameter trajectories that parameterize the gait cycle \cite{Best2023, Cortino2023, Welker2023, Huang2022, hong2023}.} Best et al.~developed a hybrid controller that uses variable impedance during stance and kinematic objectives during swing \cite{Best2023} to accurately emulate normative able-bodied kinematics and kinetics during walking over different inclines and speeds. This continuous paradigm used a parametric model to define the joint impedance, but the parameters that define the model are not necessarily interpretable or tunable.

Our previous analysis of able-bodied datasets identified consistent trends in individuality across variations of walking \cite{Reznick2020} and stair climbing \cite{Reznick2023}, presenting an interesting avenue to personalize a wide number of tasks given a few clinically accessible observations. Similar underlying trends have been validated by other groups \cite{Sharma2023, Huang2022}. 
This method isolates joint-level kinematic individuality (e.g., the difference from the population mean) at level ground and uses that baseline individuality to predict individual kinematics at other speeds and inclines, allowing efficient tuning of continuous-task kinematic controllers like \cite{Best:IROS2021}. To extend this concept to joint impedance controllers, we generalized \textit{kinetic} individuality during walking following the kinematic individuality protocol from \cite{Reznick2023}. Given this validation, we then develop a method using baseline individuality to efficiently personalize both the kinematic and impedance controllers defined in \cite{Best2023, Welker2023}.

Building on a previous tuning interface for a continuous kinematic controller for level-ground walking \cite{Quintero2018_CCI}, this study develops the technical methods and clinical interface to personalize hybrid kinematic-impedance controllers \cite{Best2023, Welker2023, hong2023} over variable-incline walking and sit-stand transitions. We define clinically meaningful parameters that map to appropriate changes in the complex, uninterpretable parameters of the continuous-phase/task models of joint kinematics or kinetics. The clinically meaningful parameters are implemented into a novel clinical tuning interface (CTI), which clinicians can use to tune level-ground walking and sit-stand transitions. Moreover, the interface automatically distributes level-ground walking individuality across continuous variations of walking, enabling a wide range of personalized behaviors on the powered prosthesis without additional configuration time. To demonstrate the tuning framework, an above-knee amputee participant was enrolled in a case study where a prosthetist used the CTI to iteratively tune the prosthesis using their clinical experience and patient feedback.

The paper is organized as follows. In Section \ref{Sec:5Methods}, we describe how individualization methods can be applied to the control models, \textcolor{black}{introduce the CTI and its tunable parameters, and describe the experimental protocol (Figure~\ref{fig:flowchart}).} In Section \ref{Sec:5Results}, we present the results of the case study with an above-knee amputee participant, assessing the efficacy of the tuning method to individualize the controller. Section \ref{sec:newConclusion} concludes with a discussion of the interactions with our clinical partner and their perception of the tuning process.

\section{Methodology}\label{Sec:5Methods}
\subsection*{Quantifying and Distributing Individuality Across Tasks}
\label{Sec:3dataset}
\textcolor{black}{To facilitate efficient individualization of joint torques in a clinical setting, we implemented a strategy to quantify and distribute an individual's kinematic/kinetic contribution (IKC) across tasks, using only individual characteristics from one task (i.e., one discrete speed and incline combination) to individualize the entire range of tasks.} We performed an offline validation of kinetic prediction using the kinematic analysis methods of \cite{Reznick2023} in Appendix \ref{Appendix:IKC}. This validation utilizes the dataset in \cite{Reznick2021,ReznickFigshare} which reports lower-limb kinematics and kinetics of ten able-bodied participants walking at multiple inclines (±0\textdegree, 5\textdegree, and 10\textdegree) and speeds (0.8, 1, 1.2 m/s), amongst other activities.

\begin{figure*}[h]
    \centering
    \includegraphics[width=0.9\textwidth]{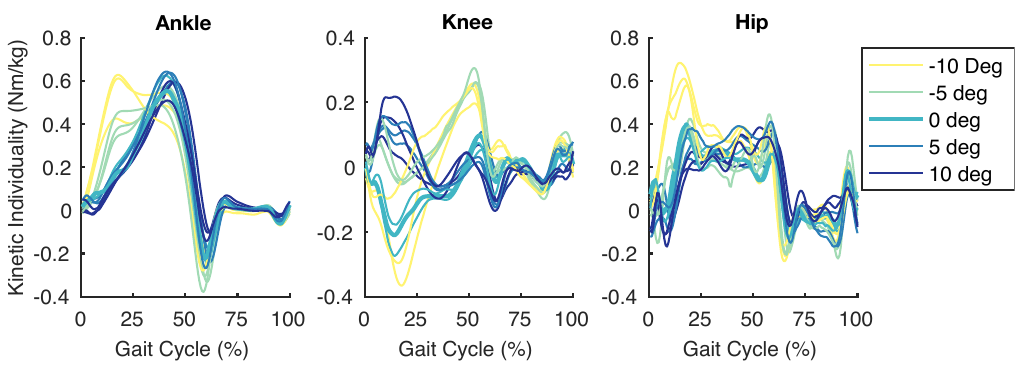}\vspace{-3mm}
    \caption{Individual Kinetic Contributions for one subject across all walking tasks within each joint. The bolded trajectory in each figure is the representative task used as the walking baseline. Note the similarities and differences between inclines.}
    \label{fig:IKC_kinetic}
\end{figure*}

We define the kinematic or kinetic IKC for a given joint and task as the difference between the individual's joint data ($\humandata$) and the leave-one-out average (LOO; $\mean{\humandata}$) of the other nine subjects. In mathematical terms, the IKC is calculated by
\begin{equation}
\label{eq:IKC-def}
\contribution_{\phase, \task, \subject} = \humandata_{\phase, \task, \subject} - \mean{\humandata}_{\phase, \task}, 
\end{equation}
for gait phase $\phase=1,\ldots,150$, task $\task=1,\ldots,N$, and subject $\subject = {1, \ldots, 10}$, where $N$ is the number of task variations in the dataset. Figure~\ref{fig:IKC_kinetic} shows the kinetic IKC trajectories for one subject.

The objective of our approach is to individualize variations of walking (e.g., different speeds/inclines) using a minimal amount of subject-specific data. Therefore, we only use individuality from level ground movement (i.e., baseline, $\task_B$) to predictively individualize non-baseline tasks \textcolor{black}{(specifically inclines and declines)} by adding the calculated IKC \textcolor{black}{(kinetic during stance and kinematic during swing)} to the population average \textcolor{black}{(see flowchart in Figure~\ref{fig:flowchart})}. Appendix~\ref{Appendix:IKC} shows that the \emph{baseline} kinetic individuality (IKC calculated at level-ground task $\task_B$) provides a good estimate of kinetic individuality across other walking tasks.

\subsection*{Walking Controller}
The tuning interface is designed to allow the clinician to modify the assistance provided by the powered leg to suit an individual user. These tuning methods are achieved by either modifying the optimization constraints used to generate the control model, or by modifying the training data for each model. This section briefly describes each of the controllers and how those models are adjusted.

For this study, we individualized the data-driven continuous-phase/task controller developed by Best et al.~\cite{Best2023}. This hybrid controller utilizes impedance control during stance and kinematic control during swing. The stance impedance controller is defined by functional representations of impedance parameters $K$, $B$, and $\theta_{eq}$. In swing, the angular position references of the knee and ankle are determined by the continuous kinematic model developed by Embry et al.~\cite{Embry2018}. These piecewise controllers are both parameterized as continuous functions of gait phase $s$ and task $\chi = (\nu, \gamma)$, where $\nu$ is the subject’s walking speed and $\gamma$ is the ground slope. Both controllers are trained on data from Section \ref{Sec:3dataset}\cite{Reznick2021}. We present simplified explanations of each controller here; further details can be found in the respective papers. 
\subsubsection*{Stance—Impedance} \label{Subsec: 5Stance-Imp}
Best et al.~\cite{Best2023} use 4th order polynomial functions ($p=4$) to model joint impedance parameter trajectories at each task. These parameters are defined as linear combinations of those phase-varying polynomials: 
\begin{equation} \label{eq: impParam}
    K_\chi = \sum_{i=0}^p k_i s^i, \ \ \theta_{eq,\chi} = \sum_{i=0}^p e_i s^i, \ \ B_\chi = \sum_{i=0}^p b_i s^i
\end{equation}
where $\kappa_\chi = \{(k_i, e_i, b_i) | i \in {0,...,p}\}$ are coefficients defined for each task sample. Using these coefficients, we construct an optimization problem to select the best coefficients ($\kappa_\chi^*$) to represent joint torques, $\tau$, from\cite{Reznick2020},
\begin{equation*}
    \kappa_\chi^* = \arg \min \frac{1}{n}\|\tau - \hat{\tau}\|,
\end{equation*}
\begin{equation}  
    \mathrm{where} \hspace{2mm} \hat{\tau} = -K_\chi(s)(\theta - \theta_{eq,\chi}(s)) - B_\chi(s)\dot{\theta}.
\end{equation}
This optimization is constrained to ensure that the solution remains within ranges that are both physiologically realistic and feasible for the prosthesis by enforcing minimum stiffness and damping and maximum damping values. These constraints are represented and enforced as vectors, and allow for changing values over the gait cycle with a unique minimum stiffness at heel strike. This optimization problem results in task-specific trajectories for $K, \theta_{eq},$ and $B$ that allow us to replicate normative able-bodied joint torques during stance. In practice, this controller is solved for each subject, task, and joint.
\subsubsection*{Swing—Kinematic}
Embry et al.~\cite{Embry2018} predictively modeled able-bodied joint kinematics with training data from \cite{Reznick2021} comprising 10 subjects at 15 tasks. These kinematics are modeled as the weighted summation of $N$ basis functions of phase, $b_k(s)$. The weight of each basis function changes for each task, as determined by the task functions $c_k(\chi)$. This gives the following expression for the desired joint angle $\theta^d$ of the knee or ankle:
\begin{equation}
    \theta^d (s,\chi) = \sum_{k=1}^N b_k(s)c_k(\chi),
\end{equation}
for $k \in 0,...,N$. The basis functions are parameterized as finite Fourier series of degree 10 to ensure periodicity, and the task functions are modeled as 2nd or 3rd degree Bernstein basis polynomials, depending on joint. Together, these basis and task functions create the kinematic model $\theta^d (s,\chi)$ parameterizing how gait cycle phase, speed, and slope affect the joint kinematics.
\subsection*{Sit-Stand Controller}
The sit-stand controller follows the same optimization paradigm as the walking controller (Section \ref{Subsec: 5Stance-Imp}), where the continuous stiffness, equilibrium angle, and damping parameters are solved from able-bodied joint torques using fourth-order polynomials. This method, originally developed by Welker et al.~\cite{Welker2023}, uses the global thigh angle to parameterize phase from sitting ($s = 0$) to standing ($s = 1$), and from 1 back to 0 for stand-to-sit. While this parameterization neatly organizes phase, the differences in normative dynamics between sitting and standing motions makes it difficult for the user to initiate stand-to-sit in practice. This is resolved by decoupling the sit-to-stand and stand-to-sit equilibrium angles at the knee, adding a dependency on joint angle velocity. This can be seen in the partial reformulation of (\ref{eq: impParam}):
\begin{equation*}
  K = \sum_{i=0}^p k_i s^i,   \theta_{eq,\dot{\theta}} = \sum_{i=0}^p e_i(\dot{\theta}) s^i, B = \sum_{i=0}^p b_i s^i
\end{equation*}  
\begin{equation}
    \mathrm{where} \hspace{2mm} e(\dot{\theta}) = e_1 + f(\dot{\theta})e_2.
\end{equation}
In this formulation, $e_1$ and $e_2$ are two independent trajectories comprising equilibrium angle, and $f(\dot{\theta}) \in [0, 1]$ is a saturating ReLU function that is active for the knee joint. While further details can be found in \cite{Welker2023}, this reformulation redefines $\kappa = \{(k_i, b_i, e_{1,i} e_{2,i}) | i \in {0,...,p}\}$ in the optimization problem
\begin{equation*}
    \kappa^* = \arg \min \|\tau - \hat{\tau}\|,
\end{equation*}
\begin{equation}
    \mathrm{where} \hspace{2mm} \hat{\tau} = -K(s)(\theta - \theta_{eq}(s,\dot{\theta})) - B(s)\dot{\theta}.
\end{equation}

\subsection*{Clinical Tuning Interface}
\begin{figure*}
\centering
\subfloat[]{\includegraphics[width=0.5\textwidth]{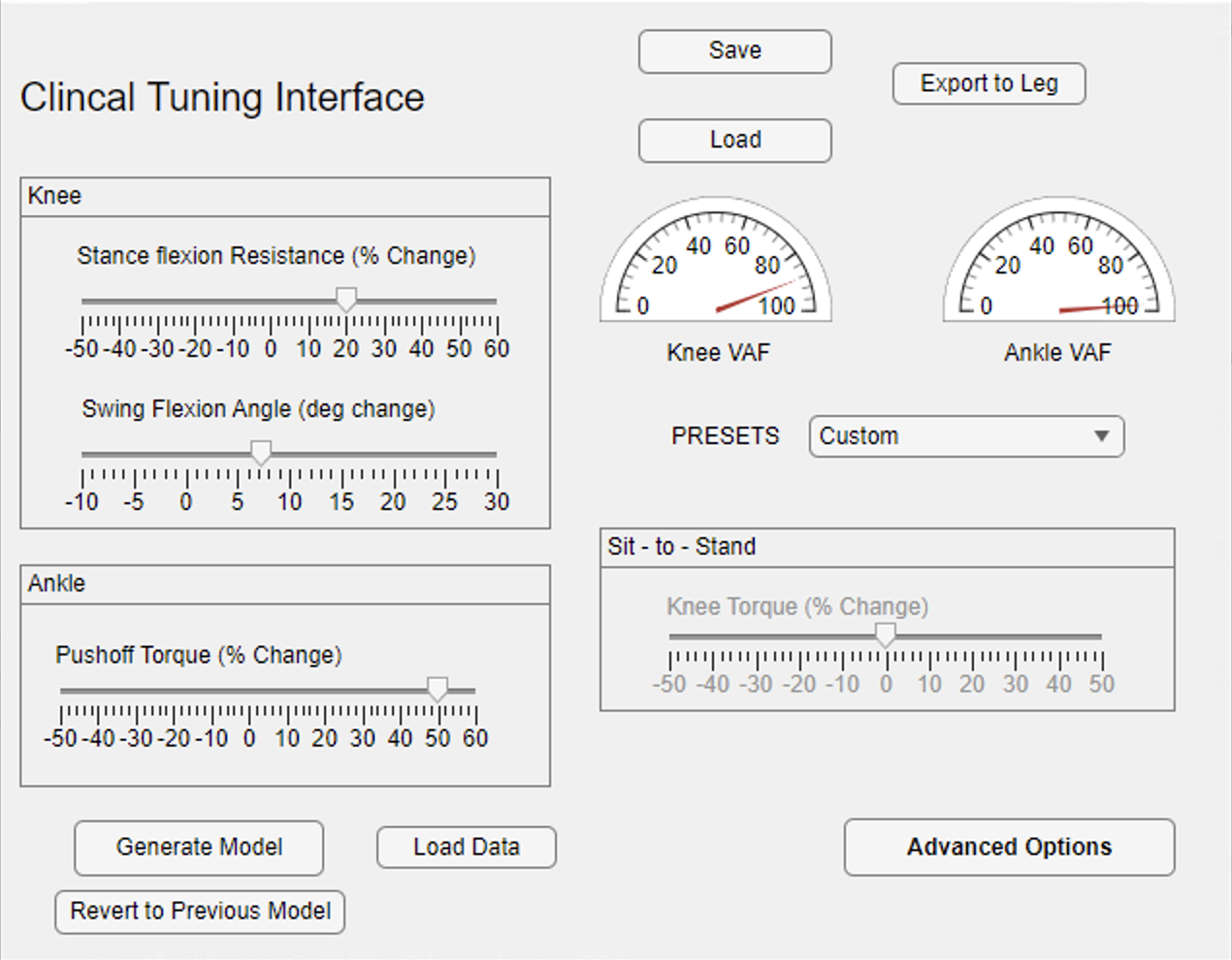}\label{fig:5-GUI}}
\hspace{.5in}
\subfloat[]{\includegraphics[width=0.35\textwidth]{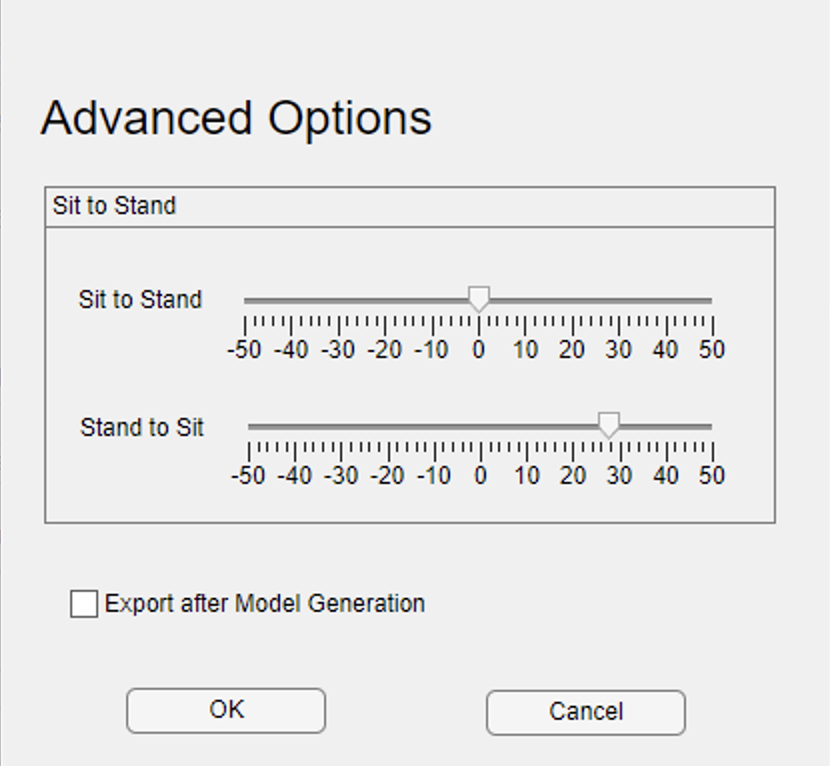}\label{fig:5ADV}}
\caption{\textcolor{black}{(a) The} CTI with adjustable parameters. A dropdown menu allows the clinician to quickly select preset profiles. Saving and loading of a profile can be completed instantaneously, and this functionality was used to switch between tuned and untuned profiles for testing. Generate Model generates only the models with parameter changes to process the models quickly. \textcolor{black}{Variance Accounted For (VAF)} gauges for each joint allow for immediate knowledge of fit. Export to Model writes the new model to the board on the prosthesis, which can be updated instantly by an engineer when prompted. \textcolor{black}{(b) The} advanced options dialog box, where the clinician can separate out sit-to-stand and stand-to-sit support.}
\label{fig:5GUI}
\end{figure*}

\textcolor{black}{The CTI (Figure~\ref{fig:5GUI}) implements tuning methods that require minimal training and clinical equipment through the careful implementation of meaningful tuning parameters for level-ground walking and sit-stand.} These parameters are similar to those of common commercial devices, and automatically translate to changes in the existing model parameters of the controller. \textcolor{black}{Because the controllers are designed to provide able-bodied torques to the user, the tunable parameters are expressed as differences from normative values in either percentage or degrees (e.g., 50\% more push off torque at the ankle as seen in Figure~\ref{fig:5GUI}a).} The allowed range for each parameter was chosen heuristically, with the bounds set at a safe but extreme value. \textcolor{black}{Figure \ref{fig:5GUI} also shows several usability features, such as preset profiles, easy saving and loading, and gauges that display Variance Accounted For (VAF), a normalized metric of model fit that is similar to the model's $R^2$.} We now discuss the implementation of each tunable parameter in the context of the model re-optimization. \textcolor{
black}{These tunable parameters define the user's kinematic or kinetic IKC, represented by $\psi$, which is automatically applied to the incline and decline walking controllers as previously described.}

\subsubsection*{Walking - Stance Flexion Resistance}
In commercial devices, stance flexion resistance determines how much force it takes to flex the knee at heel strike; this is often changed in passive devices to minimize foot slap \cite{Berger2012}. In this powered device, foot slap is not an issue because we actively control the ankle, so the parameter effectively changes how rigid the knee is as the user progresses through weight acceptance in double support. This change is implemented by changing the heel strike constraint (Section \ref{Subsec: 5Stance-Imp}), which is bounded to [-50\%, 60\%] of the published value. Anecdotally, this was often described as how `squishy' the knee felt at heel strike. 

\subsubsection*{Walking - Swing Knee Flexion}
This parameter allows the clinician to tune the swing knee kinematics across all walking tasks in degrees (with bounds [-10, 30]) using clinical descriptions and units. The training trajectories of the knee model optimization (from able-bodied data \cite{Reznick2021}) are modified through the addition of a spline ($\psi_{Flx}$) that smoothly applies changes across phase. The default spline is set to zero and upon tuning increases flexion smoothly around peak swing knee flexion to the tuned parameter. This spline is added uniformly to the training trajectories for all walking tasks, given that we previously observed that kinematic individuality at level ground is similar to all other variations of walking \cite{Reznick2021, Reznick2023}.

\subsubsection*{Walking - Push-Off Torque} 
This parameter scales the reference ankle torque ($\tau_{ref}$, again from able-bodied data \cite{Reznick2021}) used to train the impedance model during the push-off portion of stance. We leverage the generalizable trends of joint-wise kinetic individuality by smoothly scaling the ankle torques around the push-off peak of the baseline task (level ground, $\chi_B$) using a spline. This spline ($\gamma_{Pot}$) is used to specifically scale peak torque, while leaving other parts of the gait cycle unchanged. An untuned spline is equal to 1 at all points, and tuning the spline modifies the scaling factor specifically around push off. The parameter is bounded between $[-50\%, 60\%]$, resulting in a spline with peak values of -0.5 Nm/kg or -1.6 Nm/kg, respectively. To apply this tuning to the new reference torque $\tau_\chi$, the spline scales the level-ground joint torques $\tau_{ref,\chi_B}$, the scaled difference is isolated, and then that level-ground individuality $\psi_{Pot}$ is added to all tasks: 
\begin{eqnarray}
    \psi_{Pot} &=& (\tau_{ref,\chi_B} \cdot \gamma_{Pot}) -\tau_{ref,\chi_B},\nonumber\\
    \tau_\chi &=& \tau_{ref,\chi} + \psi_{Pot}.
\end{eqnarray}
To make the process more accessible to the clinical team, we use a scaling percentage instead of using the units of Nm/kg to describe the change.
\subsubsection*{Sit-to-Stand - Knee Torque}
Tuning knee torque during sit-stand transitions is achieved by scaling the reference torque profiles input into the optimization problem. \textcolor{black}{The scaling remains constant throughout each motion; however, sit-to-stand and stand-to-sit motions were tuned independently (Figure~\ref{fig:5GUI}b).} Due to the decoupling of the training parameters during tuning and the equilibrium angles during optimization, there were large changes in equilibrium angles while the stiffness and damping values remained relatively stable across different parameter selections. With split tuning, small variations in thigh velocity can lead to abrupt switching between controllers during the stand-to-sit transition, particularly when the standing and sitting parameters differ substantially. To fix this, we linearly interpolate the scaling factor during the last 10\% of standing and the first 10\% of sitting. This fix is also applied during sitting when the separation between the sit-to-stand and stand-to-sit parameters is greater than 60\%, consequently, we constrained the separation to 60\%.  

\subsection*{Experimental Methods}
The experimental protocol was approved by the Institutional Review Board of the University of Michigan (HUM00166976). 

\subsubsection*{Subject Preparation}
For this case study, the subject was an above-knee amputee participant who had previously walked on the baseline controller. The participant (TF01: Male, 27yrs, 116kg) was fit with the powered knee-ankle prosthesis by a clinician, who adjusted the pylon length and alignment for optimal fit. There was an existing relationship between the participant and clinician, allowing for a high level of communication during the study protocol. Reflective markers were placed on anatomical landmarks following the Plug-in-Gait marker set \cite{Leboeuf2019} with additional markers at the medial knee, medial ankle, and greater trochanter for post-processing. 

\subsubsection*{Tuning Protocol}
The tuning session was designed to mimic a typical session in a clinic, and began with the user walking on a level, 7~m handrail walkway with the baseline controller until comfortable not using the handrails. Then, the clinician cycled through a series of preassigned profiles which showcased a high and low setting for each parameter, assessing TF01's preference. These baselines were designed to apply values at $\pm 80\%$ of the parameter space (e.g., a parameter that allows up to a $\pm50\%$ change has a HIGH preset of $+40\%$) to calibrate both the user's and prosthetist's mental models. 

Then, with the initial preferences in mind, the prosthetist iteratively tuned the prosthesis. This communication was similar to an optometrist visit, where the subject was asked to choose between two options until both the user and clinician agreed that the gait was comfortable. The prosthetist tuned the walking controller, then the sit-stand controller. \textcolor{black}{The manually-tuned IKC from the level-ground walking controller was automatically applied the incline and decline walking controllers.}

\subsubsection*{\textcolor{black}{Assessment Protocol}}
This protocol was run on both the tuned and un-tuned profiles to determine preference. The order of tested profiles was blinded and randomized, and after each test, the subject was asked to choose their preferred condition. The user performs three trials of a 5-time sit-to-stand test (5xSTS), for 15 total sit-stand repetitions. \textcolor{black}{Then, the subject walked on an instrumented treadmill (Bertec, Columbus OH) for 30 seconds at -10$^{\circ}$, 0$^{\circ}$, and 10$^{\circ}$, where the incline conditions tested the automatically tuned controllers based on level-ground tuning parameters (IKC). While overground walking trials occurred, they were omitted from these results because they did not contribute a large enough number of steady-state strides at consistent speeds.}

\section{Results} \label{Sec:5Results}

\begin{figure}[h]
    \centering
    \includegraphics[width=.8\linewidth, ]{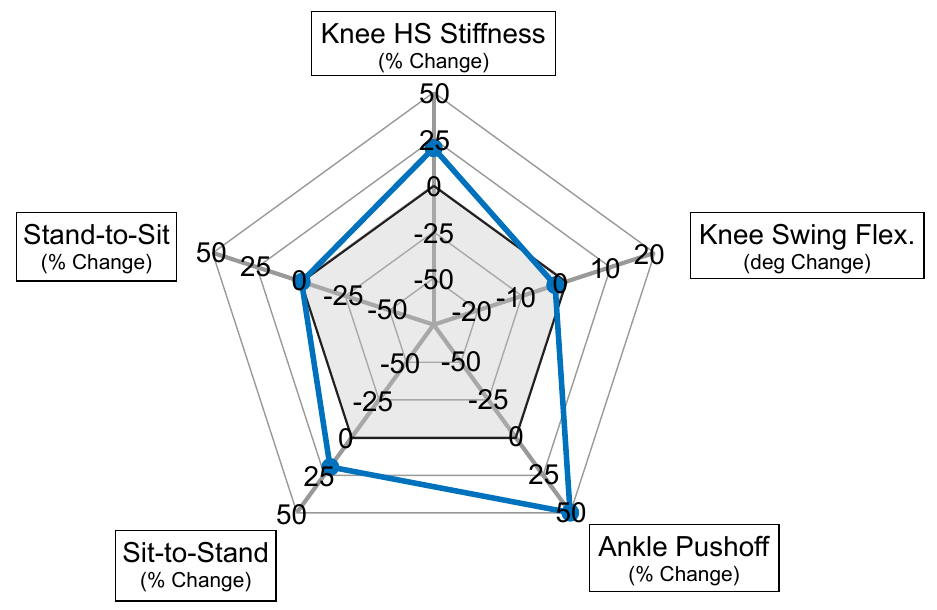}
    \caption{Final individualized parameters for TF01 given the scale displayed in the UI. The shaded region covers values below normative able-bodied parameters.}
    \label{fig:5TuningParam}
\end{figure}

The subject's prosthesis was tuned by their prosthetist until both parties were happy with the tuned profile. Each tuning iteration of the walking controller took $2\pm 0.1$ minutes, with 8 iterations. The sit-stand controller went through 5 iterations and took $1\pm 0$ minutes on average. The final tuned configuration for the subject can be seen in \autoref{fig:5TuningParam}. No changes were made to swing knee flexion or stand-to-sit torque, moderate changes were made to sit-to-stand torque and knee stiffness, and the ankle push-off was increased by 50\%. No parameters were selected to provide less assistance than the baseline controller.
\begin{figure}[t]
    \centering
    \includegraphics[width=.9\columnwidth]{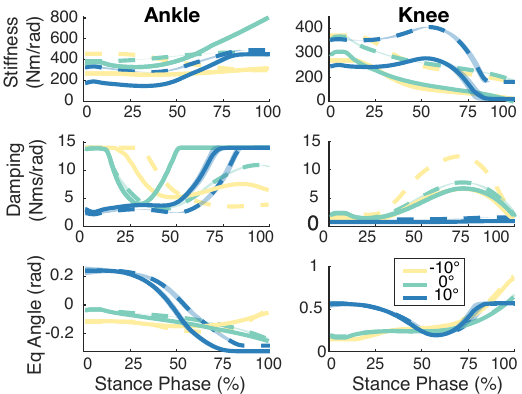}
    \caption{Continuously varying impedance parameters across different inclines for the ankle (left) and knee (right). The solid lines represent the untuned baseline configuration, while the dashed are individualized.}
    \label{fig:5ImpParam}
\end{figure}

\begin{figure}[t]
    \centering
    \includegraphics[width=.8\columnwidth]{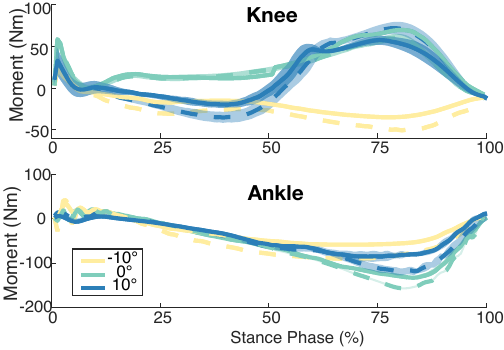}
    \caption{Prosthetic knee (top) and ankle (bottom) torques at each incline in the baseline (solid) and individualized (dashed) controllers. The shaded regions indicate standard deviation across all strides.}
    \label{fig:5ankTor}
\end{figure}

\begin{figure}[t]
    \centering
    \includegraphics[width=.9\columnwidth]{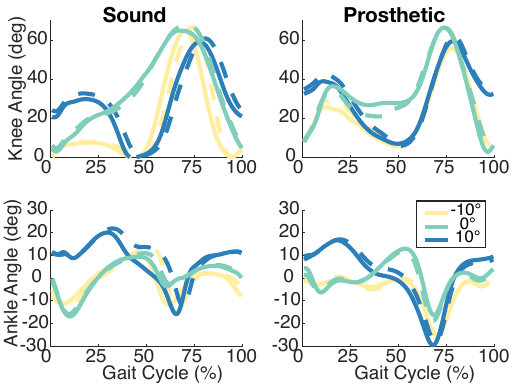}
    \caption{Knee (top) and ankle (bottom) angles from the sound (left) and prosthetic (right) sides at each incline for the baseline (solid) and individualized (dashed) controllers.}
    \label{fig:5KneeAng}
\end{figure}

\begin{figure}[t]
    \centering
    \includegraphics[width=0.9\columnwidth]{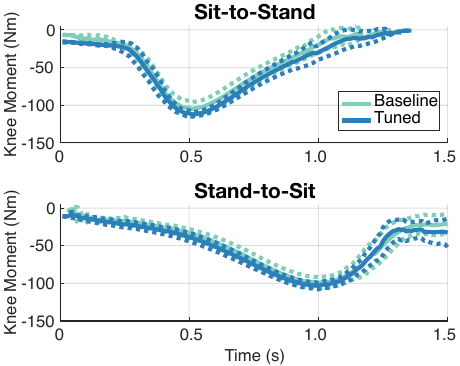}
    \caption{Prosthetic knee torques for Sit-to-Stand (top) and Stand-to-Sit (bottom) with the tuned and baseline controllers. Solid lines represent the mean and dashed lines represent $\pm$ SD.}
    \label{fig:5sts}
\end{figure}

\autoref{fig:5ImpParam} shows how tuning knee stiffness at heel strike affects the impedance parameter trajectories over stance phase. Heel strike stiffness increased 25\% over the untuned baseline controller. This parameter changes the lower bound of the optimized stiffness to ensure a more rigid knee at foot contact. This inequality constraint ensures that a less rigid solution is not allowed, but if the solution falls above the limit, like we see in the tuned controllers, some variance is expected. 

Despite the fact that the tuned constraint is only enforced at initial foot contact, the stiffness stays fairly constant in early stance instead of decreasing with the constraint. This behavior is echoed by the slightly higher torques during early stance of inclined and declined walking, see \autoref{fig:5ankTor}.

Next, we look at the ankle push-off tuning in the CTI. \autoref{fig:5ankTor} shows that while a +50\% change was applied in the tuning parameter, the achieved peak torques increased closer to 25\% across tasks (20\%, 24\%, and 29\% at decline, level, and inclined tasks). Each incline shows meaningful increases, but not of the magnitude expected. Looking at the impedance parameters in \autoref{fig:5ImpParam}, we can see that the optimization accommodated the request for additional ankle pushoff by increasing stiffness throughout the gait cycle and increasing damping in late stance. The prosthetic ankle kinematics in \autoref{fig:5KneeAng} exhibit a slight shift in phase around the plantarflexion peak, though no major changes in magnitude can be observed from the added push-off torque. On the sound side, there is a significant shift in both magnitude and phase of motion during late stance/early swing (25-75\%). Similar changes in phase are observed in declined and level walking, though not as extreme as inclined walking. The change in shape around 40\% gait is more similar to the prosthetic side, indicating increased inter-limb symmetry. 

\autoref{fig:5KneeAng} also reveals changes in the shape of the first peak of prosthetic knee flexion that corresponds with loading a stiffer knee. On the sound side, we observe increases in knee flexion at all inclines, as well as a shift in phase of both peak knee flexion and extension that aligns better with the prosthetic side, indicating a more symmetrical gait. 

Tuning the sit-stand controller worked as expected when tuning the knee assistance (see \autoref{fig:5sts}). The increased support requested via the CTI significantly increased peak torques during sit-to-stand; the prescribed 20\% increase resulted in a 7.4\% increase in peak torque, with no significant changes in the motion pattern. Further, there was no significant change in peak moment at the knee in the stand-to-sit motion, as there was no change from the baseline. This motion was recorded using a 5xSTS test, a temporal clinical test. In the baseline condition, it took 15.75 $\pm$ 0.49 seconds to complete the test, while the individualized condition took 14.66 $\pm$ 0.79 seconds. The subject performed the task 1.09 seconds faster on average, and the subject preferred the individualized support. 

\section{Discussion and Conclusion}
\label{sec:newConclusion}
TF01's extensive experience with the un-tuned controllers led to mixed preferences. The subject selected significant changes to the controller and could easily perceive the differences during tuning sessions. However, when blinded to conditions, TF01 preferred the baseline walking controller. \textcolor{black}{We believe that this is largely based on the subject's prior experience with the baseline controller and inadequate acclimation time with the tuned controller.} For sit-to-stand transitions, TF01 favored increased assistance when rising from a seated position but chose lower sitting torque to enable faster motion. The parameter distinction between sitting and standing was crucial for successful controller tuning. We observed that more extreme modifications led to the exclusion of trajectories from the training set, indicating increased difficulty in solving the problem. Nonetheless, the accepted models maintained errors below 5\% at the ankle and 15\% at the knee, effectively reducing commanded peak values for tasks. 

Feedback from clinicians and subjects indicated noticeable effects of selected parameters on the controller, with torque changes more apparent to wearers and kinematic changes more evident to clinicians. Despite the CTI offering a curated selection of tunable parameters, clinicians were able to fully customize behavior to suit the wearer's needs. Knee stiffness elicited strong opinions, with users immediately preferring a stiffer knee after trying pre-designed profiles. Swing knee flexion unexpectedly served a dual purpose, influencing shank progression dynamics in addition to the intended effect of ensuring toe clearance. Ankle push-off was increased to prevent a sensation of falling into the next stride, but fine-tuning was necessary to prevent excessive torque that could destabilize users. Adjustments in this parameter often required instructing users to load the prosthetic side more fully.

While each tuning parameter was intended to affect a specific aspect of the controller, the complexity of gait means changes in one parameter are intricately linked to others. We observed that commanded torque changes were not always fully realized, potentially due to inappropriate utilization by subjects. For instance, if ankle push-off torque doesn't propel the subject's center of mass (COM) forward, the torque applied ends up effectively pushing the limb around without causing larger changes in the COM. In this case, the extra push-off kicked the leg into a higher knee flexion during inclined and declined walking on the prosthetic side, causing higher knee flexion during swing at those joints, despite no additional flexion being commanded from the CTI. These interrelated effects are an interesting avenue for potential research, but go beyond the scope of this paper, which is intended to present an interface. This mismatch underscores the importance of involving clinical partners who consider all aspects of gait in the tuning process. Sit-stand personalization was more straightforward, with significant changes in peak joint moment during sit-to-stand motions reflecting clinician intentions.

There are a few points that we found interesting in this work. Firstly, the baseline controller is designed to emulate normative able-bodied motion and the subject chose strong parameter changes during tuning, but during walking trials TF01 indicated  a preference for the normative, biomimetic motion. Secondly, knee swing flexion had complex effects on gait, balancing toe clearance with shank progression speed for comfortable socket-limb dynamics. Lastly, tuning specific parameters led to cascading effects on gait, especially as tuning diverged from normative values. \textcolor{black}{These findings highlight potential directions for future research and the development of automated tools to enhance the efficiency and effectiveness of the tuning process.}

There are some limitations to this study that have potential implications. First, the clinician indicated that the next step after preliminary tuning would be to send the wearer home and have them return in a few weeks for tuning refinement, which is typical for a new prosthesis but not possible with our research hardware. Next, a larger subject sample size is needed to understand the range of possible parameter changes in the population. Lastly, observational gait analysis is notoriously subjective, and tuning results are dependent on both the clinician and subject. 

\vspace*{0.5 cm}

\section*{Acknowledgment}
This work was supported by the National Institute of Child Health \& Human Development of the NIH under Award Number R01HD09477. The content is solely the responsibility of the authors and does not necessarily represent the official views of the NIH.

\appendix
\textcolor{black}{This appendix details the validation of the individual kinetic contribution (IKC, Eq. \ref{eq:IKC-def}). The IKC of a representative subject across modes and tasks are shown in \autoref{fig:IKC_kinetic}.}

\label{Appendix:IKC}

\subsection*{Individualization of Walking Kinetics}
The objective is to individualize variations of walking (e.g., different speeds/inclines) using a minimal amount of subject-specific data. Therefore, we only use individuality from level ground movement (i.e., baseline, $\task_B$) to predictively individualize non-baseline tasks by adding the calculated IKC to the population average. Here, we test and expand the assumption of kinematic individuality made in \cite{Reznick2020}: that the \emph{baseline} (IKC calculated at $\task_B$) provides a good estimate of individuality across all walking tasks. 

We discuss error using the root mean squared error (RMSE). This metric calculates the deviation of the predicted IKC $\contribution_{\phase, \task_B, \subject}$ (with respect to baseline $\task_B$) from the subject's observed IKC $\contribution_{\phase, \task, \subject}$ by
\begin{equation}\label{eq:RMSE}
    \mathrm{RMSE}_{\task, \subject} = \sqrt{\sum_{\phase=1}^{I}(\contribution_{\phase, \task_B, \subject} - {\contribution}_{\phase, \task, \subject})^2/I},
\end{equation}
for $I = 150$ points in phase. Though we explored IKC fit at specific points along the gait cycle in \cite{Reznick2020}, here we examine the average error across the gait cycle to facilitate comparisons between methods. It should be noted that this single measure cannot differentiate between errors caused by large, brief residuals (e.g., due to a phase shift in swing knee flexion) vs. small persistent differences over the gait cycle. The RMSE is reported in Nm/kg, and the results should be interpreted within the context of the respective joint.

As a benchmark, we compare the RMSE from each individualized task to the `un-tuned' RMSE ($\contribution_{\phase, \task_B, \subject}=0$ in \autoref{eq:RMSE}), which corresponds with no individualization from the population average. Because our input dataset contained a relatively large number of representative strides collected on a treadmill (about 30 strides over 30 seconds), these models used half of the strides to calculate the baseline IKC and the other half for prediction validation to prevent overfitting. We define an improvement as a reduction in the individualized RMSE from the non-individualized RMSE, which is quantified by the difference in RMSE (positive values correspond with improvement). Kinetics are normalized by subject mass and reported in Nm/kg.

We analyze the RMSE between the predicted and observed individuality with an N-way ANOVA (MATLAB 2021a, Mathworks, Natick, MA) considering the factors of subject, velocity, incline, and joint. Then, post-hoc multiple comparison tests (Tukey-Kramer) investigate the effect of individualization methods within specific variable groups.

\subsection*{Kinetic Individuality Validation}
\begin{figure}[h]
    \centering
    \includegraphics[width=0.45\textwidth]{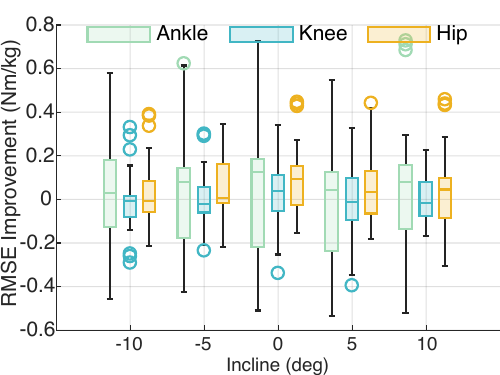}\vspace{-2mm}
    \caption{Kinetic RMSE improvement (positive) over LOO average for walking tasks at 1.0m/s for the ankle (green), knee (blue), and hip (orange). The boxes show the 25-75\% interquartile range, with a line at the median; the whiskers show the non-outlier maximums and circles are outliers. Each box represents the data across all subjects.}
    \label{fig:3KineticBar}
\end{figure}

The results of the statistical validation of kinetic individuality based on the offline dataset data are as follows.
The baseline method was more predictive than the LOO for 76\% of the all subjects and joints, showing significant improvements upon individualization for each joint across all tasks. There was a 44\% (0.085 Nm/kg) improvement in fit at the ankle,  18\% (0.024 Nm/kg) at the knee, and 31\% (0.05 Nm/kg) at the hip (\autoref{fig:3KineticBar}). Each of the joints shows a significantly improved fit upon individualization with initial one-tailed paired t-test values of $p=0.002$, $p=0.011$, and $p=0.003$ for the ankle, knee, and hip, respectively. Subject and joint specific details are given in \autoref{tab:3KineticTraj}.

The n-way ANOVA indicates that subject, incline, joint, and individualization method all significantly affect the results ($p \ll 0.05$), but velocity had no effect on the resulting RMSE. There is a large range of RMSE changes in this data study, but the results suggest that level-ground kinetic individuality is a decent, though imperfect, estimation of kinetic individuality at other tasks.

\begin{table*}[t!]
\caption{RMSE of each predicted trajectory when compared to the experimental values, similar to the kinematic calculation seen in (\ref{eq:RMSE}). Significance between walking and LOO values is represented with `*' for $p < 0.05$. These values are the average error across the gait cycle in Nm/kg.}
\resizebox{\textwidth}{!}{
\begin{tabular}{lllllllllllll}                        
&                            & AB01  & AB02  & AB03  & AB04  & AB05  & AB06  & AB07  & AB08  & AB09  & AB10  & Average                    \\ \hline
\multicolumn{1}{|l}{\multirow{3}{*}{\begin{tabular}[c]{@{}l@{}}Walking:\\WALK\end{tabular}}} & \multicolumn{1}{l|}{Ankle} & 0.091 & 0.088 & 0.120 & 0.141 & 0.142 & 0.098 & 0.064 & 0.101 & 0.121 & 0.077 & \multicolumn{1}{l|}{0.104*} \\
\multicolumn{1}{|l}{}                                                                                & \multicolumn{1}{l|}{Knee}  & 0.114 & 0.102 & 0.135 & 0.093 & 0.108 & 0.082 & 0.080 & 0.129 & 0.127 & 0.106 & \multicolumn{1}{l|}{0.108*} \\
\multicolumn{1}{|l}{}                                                                                & \multicolumn{1}{l|}{Hip}   & 0.142 & 0.098 & 0.158 & 0.089 & 0.140 & 0.107 & 0.080 & 0.102 & 0.125 & 0.091 & \multicolumn{1}{l|}{0.113*} \\ \hline
\multicolumn{1}{|l}{\multirow{3}{*}{\begin{tabular}[c]{@{}l@{}}Walking:\\ LOO\end{tabular}}} & \multicolumn{1}{l|}{Ankle} & 0.244 & 0.115 & 0.199 & 0.224 & 0.299 & 0.125 & 0.081 & 0.112 & 0.210 & 0.279 & \multicolumn{1}{l|}{0.189} \\
\multicolumn{1}{|l}{}                                                                                & \multicolumn{1}{l|}{Knee}  & 0.185 & 0.111 & 0.145 & 0.138 & 0.152 & 0.140 & 0.078 & 0.134 & 0.143 & 0.093 & \multicolumn{1}{l|}{0.132} \\
\multicolumn{1}{|l}{}                                                                                & \multicolumn{1}{l|}{Hip}   & 0.136 & 0.136 & 0.186 & 0.113 & 0.254 & 0.136 & 0.078 & 0.179 & 0.200 & 0.213 & \multicolumn{1}{l|}{0.163} \\ \hline
\end{tabular}
}

\label{tab:3KineticTraj}
\end{table*}

\section{Discussion} \label{Sec:5Discussion}
\subsection*{Kinetic Individuality Conclusions}
Our method for predicting kinetic individuality across tasks was more accurate than the LOO average alone (\autoref{fig:3KineticBar}). 
We observed more pronounced trends within subjects regarding kinetic individuality compared to those reported for kinematics in \cite{Reznick2023}, with significant improvements for all joints, when all subjects, tasks, and strides are considered (see Table \ref{tab:3KineticTraj}). 

At the joint-level, walking kinematics \cite{Reznick2023} and kinetics had a similar trend, with ankle predictions being more accurate than those for the hip on average. Upon applying our individualization method to knee kinetics, we noted its limitation in reflecting the subtle variations in flexion/extension torques associated with changes in walking incline. Despite similar challenges encountered at the hip, related to non-steady torques, our approach managed to capture prominent torque trends as depicted in Fig. \ref{fig:IKC_kinetic}.  

Walking speed was not a significant factor in the effectiveness of our method for predicting kinetic individuality, but incline was highly significant (see Fig. \ref{fig:3KineticBar}). As expected, the baseline, level-ground task showed the best individualization. The ankle showed significant improvements after individualization, likely fitting the peak torques found at \~50\% of the gait cycle, but the level-ground baseline did not account for individuality in the early stance torques. Predictions at the knee showed moderate success in tasks other than the baseline, with modest improvements offset by slight decreases in fit. The method accurately predicted the magnitude and timing of peak torques at the hip during level and inclined walking, but this precision diminished during declined walking.

As discussed earlier, this method of propagating individuality had mixed results across subjects, indicating that baseline individuality is more predictive across tasks for some subjects than others (Table \ref{tab:3KineticTraj}). For example, AB03 shows good improvement at all joints, while AB07 sees the opposite trends. These trends across subjects demonstrate particularly accurate prediction when the LOO average kinetics are different from the population average, as observed with AB04. This is especially encouraging as we are applying this method to users who have impaired walking. 

\vspace*{0.5 cm}

\bibliographystyle{ieeetr}
\bibliography{TuningProp}
%

%
%
%
%
%
%
%
\end{document}